%% file: main.tex
\documentclass[a4paper,fleqn]{cas-dc}

\usepackage[authoryear,longnamesfirst]{natbib}
\usepackage{algorithm}
\usepackage{algpseudocode}
\usepackage{amsmath}
\usepackage{amssymb}
\usepackage{array}
\usepackage{bbding}
\usepackage{booktabs}
\usepackage{color}
\usepackage{float}
\usepackage{graphicx}
\usepackage{multicol}
\usepackage{multirow}
\usepackage{threeparttable}

\def\tsc#1{\csdef{#1}{\textsc{\lowercase{#1}}\xspace}}
\tsc{WGM}
\tsc{QE}

\begin{document}
\let\WriteBookmarks\relax
\def\floatpagepagefraction{1}
\def\textpagefraction{.001}

\shorttitle{}    

\shortauthors{}


\title [mode = title]{SAM-OCTA: Prompting Segment-Anything for OCTA Image Segmentation}




%

\author[1]{Xinrun Chen}[
    orcid=0000-0001-8982-9330
]



\ead{chenxinrun@cqu.edu.cn}


\credit{Methodology, Software, Writing}

\author[1]{Chengliang Wang}[]

\fnmark[*]
\ead{wangcl@cqu.edu.cn}


\credit{Conceptualization of this study}

\author[1]{Haojian Ning}[]

\ead{nhj@cqu.edu.cn}

\credit{Data curation and analysis}

\author[2]{Shiying Li}[]

\ead{shiying_li@126.com}

\credit{Medical advisor}

\author[2]{Mei Shen}[]

\ead{luckymay2889@163.com}

\credit{Medical advisor}

\affiliation[1]{organization={College of Computer Science, Chongqing University},
            addressline={No.174 Shazheng St., Shapingba District}, 
            city={Chongqing},
            postcode={400044}, 
            state={},
            country={China}}

\affiliation[2]{organization={Xiang’an Hospital of Xiamen University},
            addressline={No.2000, Xiang'an East Road}, 
            city={Xiamen},
            postcode={361104}, 
            state={},
            country={China}}

\cortext[1]{Corresponding author}



\maketitle

\input{sections/0_abstract}
\input{sections/1_introduction}
\input{sections/2_related_work}
\input{sections/3_method}

\input{sections/4_experiments}

\input{sections/5_discussion}

\input{sections/6_conclusion}


\section*{Declaration of Competing Interest}

The authors declare that they have no known competing financial 
interests or personal relationships that could have appeared to influence the work reported in this paper. 

\section*{Declaration of Data Availability}

The data is publicly available, and the code has been public on GitHub.

\section*{Acknowledgement}

This work is supported by the Chongqing Technology Innovation $\And$ Application Development Key Project (cstc2020jscx; dxwtBX0055; cstb2022tiad-kpx0148).

\printcredits

\bibliographystyle{cas-model2-names}

\bibliography{cas-refs} 

\bio{}
\endbio


\end{document}

%% file: sections/0_abstract.tex
\begin{abstract}
Segmenting specific targets or biomarkers is necessary to analyze optical coherence tomography angiography (OCTA) images. Previous methods typically segment all the targets in an OCTA sample, such as retinal vessels (RVs). Although these methods perform well in accuracy and precision, OCTA analyses often focusing local information within the images which has not been fulfilled. In this paper, we propose a method called SAM-OCTA for local segmentation in OCTA images. The method fine-tunes a pre-trained segment anything model (SAM) using low-rank adaptation (LoRA) and utilizes prompt points for local RVs, arteries, and veins segmentation in OCTA. To explore the effect and mechanism of prompt points, we set up global and local segmentation modes with two prompt point generation strategies, namely random selection and special annotation. Considering practical usage, we conducted extended experiments with different model scales and analyzed the model performance before and after fine-tuning besides the general segmentation task. From comprehensive experimental results with the OCTA-500 dataset, our SAM-OCTA method has achieved state-of-the-art performance in common OCTA segmentation tasks related to RV and FAZ, and it also performs accurate segmentation of artery-vein and local vessels. The code is available at https://github.com/ShellRedia/SAM-OCTA-extend. 
\end{abstract}


\begin{highlights}
\item  The fine-tuned Segment Anything model works well in OCTA segmentation.
\item  The adaptation of prompt points has been clarified in multiple segmentation tasks.
\item  An effective approach is provided for artery-vein segmentation in OCTA images.
\end{highlights}

\begin{keywords}
Optical Coherence Tomography Angiography \sep Image Segmentation \sep Model Fine-tuning \sep Prompt Point
\end{keywords}

%% file: sections/1_introduction.tex
\section{Introduction}\label{intro}

The retina has fascinated researchers and ophthalmologists due to its highly specialized nature. Optical Coherence Tomography Angiography (OCTA) is a non-invasive imaging technique that visualizes retinal microvasculature in high resolution without dye injection \citep{lains2021retinal, wang2021deep}. It provides qualitative and quantitative information about blood flow contrast in various retinal layers, making it a valuable tool for disease staging and preclinical diagnosis, including diabetic retinopathy and age-related macular degeneration \citep{liang2021foveal, lopez2023fully}.

 The specific retinal structures, such as the retinal vessels (RVs) and foveal avascular zone (FAZ), have been targeted for segmentation in research to aid in the detection and monitoring of retinal diseases \citep{hu2022joint, li2022rps}. Some computer-aided diagnosis (CAD) systems have also been developed, contributing to the clinical evaluation of various ophthalmic diseases \citep{vali2023cnv, meiburger2021automatic}. Researchers have been actively exploring deep learning-based methods for image quality assessment and segmentation with limited datasets to address these challenges and enhance the accuracy and efficiency of OCTA image analysis. 
 
Most models for OCTA segmentation have self-designed network modules and structures, often requiring training from scratch with specific datasets. Due to the generally small scale of existing OCTA public datasets, these methods are prone to overfitting.  Foundational models have revolutionized artificial intelligence (AI) with extensive pre-training on web-scale datasets and powerful generalization. In various AI areas, including computer vision and NLP, large AI models are actively researched, fostering future advancements and breakthroughs \citep{zhang2023comprehensive, shi2023generalist}.

Segment Anything Model (SAM) is a foundational model designed for image segmentation. Without additional training, the model achieves satisfactory performance in various regular segmentation tasks \citep{kirillov2023segment}.  Accurately identifying anatomical structures and lesions is crucial for an effective medical segmentation model. Accurately identifying anatomical structures and lesions is crucial for the effectiveness of medical segmentation models, and applying SAM to medical image segmentation poses a series of challenges. Medical images differ significantly from natural images in quality, noise, resolution, and other factors, impacting SAM's segmentation performance. Therefore, further optimization efforts are necessary to fully harness SAM's potential in the field of medical image segmentation \citep{cheng2023sam, zhang2023segment}.

Adopting a fine-tuning approach to SAM and introducing prompt information can enhance and guide the model's segmentation, aiming to improve some complex OCTA segmentation cases. We summarize the contributions of this paper as follows:

\begin{enumerate}[a)]

\item 
We employ SAM, a pre-trained foundational model with prompt points, for the segmentation task of OCTA images for the first time. The fine-tuned model has been tested on the OCTA-500 dataset, achieving or approaching the results at the state-of-the-art level.

\item 
We fine-tune the SAM's parameters using OCTA datasets and the LoRA (Low-Rank Adaptation) method. This process aims to adapt SAM to the segmentation task on OCTA images while preserving its solid semantic understanding of images.

\item 
We propose a prompt point generation strategy for OCTA image segmentation tasks and conduct a comprehensive experimental analysis of the effectiveness of prompt points in RV, FAZ, capillary, and artery-vein (AV) segmentation tasks.

\end{enumerate}  

This paper is an extension of our conference paper \citep{wang2023sam}. The new contributions of this paper include:

\begin{enumerate}[a)]

\item 
We provide an additional strategy for generating prompt points: selecting the bifurcations, intersections, and endpoints of RV, artery, and vein with sparse annotation. This aids in analyzing and interpreting the relationship between the placement of prompt points and segmentation performance.

\item 
We tested the segmentation tasks' performance on SAM models of different types, which helps in selecting configurations for practical applications.

\end{enumerate}  

%% file: sections/2_related_work.tex
\section{Related Work}

\subsection{OCTA Segmentation Models}

The vision transformer (ViT) is a typical model architecture for image processing tasks \citep{dosovitskiy2020image}. Its variants are frequently employed in OCTA-related tasks. TCU-Net is an efficient cross-fusion transformer to achieve continuous vessel segmentation, addressing issues like vessel discontinuities or missing segments while maintaining linear computational complexity \citep{shi2023tcu}. The OCT2Former employs a dynamic transformer encoder capturing global retinal vessel context \citep{tan2023oct2former}. StruNet combines the swin-transformer modules and the residual blocks for OCTA image denoising \citep{ma2023strunet}. VCT-NET adopts a dual-branch network design fusion of the features extracted from the swin-transformer modules and the convolutional modules for vessel segmentation \citep{liu2022vct}. 
ARP-Net is another model based on the adaptive gated axial transformer, which supports repairable vessel segmentation \citep{liu2023transformer}.

The above method is sufficient to demonstrate the effectiveness and potential of the ViT-based SAM in OCTA images. Some more supervised segmentation methods typically follow a coarse-to-fine feature extraction process. IPN is a 3D-to-2D segmentation network for OCTA images and presents a projection learning module that enables effective feature selection and dimension reduction \citep{li2020image}. \citet{ma2020rose} presents a split-based coarse-to-fine RV segmentation network for OCTA images, demonstrating robust and accurate vessel segmentation of retinal. OVS-Net adopts a two-stage cascaded structure to process OCTA images and provide continuous probability maps for vessel segmentation \citep{zhu2022ovs}. RPS-Net consists of parallel modules for learning global semantic features and incorporating local details from the volumetric data to obtain more accurate segmentation results \citep{li2022rps}. FARGO adopts a ResNeSt-based encoder with split attention for FAZ segmentation while employing a cascaded network for RV segmentation as an auxiliary task\citep{peng2021fargo}. BSDA-Net uses boundary heatmap regression and signed distance map reconstruction branches for accurate FAZ contour segmentation \citep{lin2021bsda}. DB-UNet introduces a dual-branch multi-layer perceptron (MLP) mixer segmentation technique, which is able to accurately segment RVs with low computational complexity \citep{wang2023db}.

Furthermore, the recent rise of unsupervised methods has led to some research applying relevant techniques to OCTA images. These methods can extract certain features across datasets, enabling segmentation or denoising tasks to be accomplished \citep{liang2021foveal, ma2022retinal}. The abovementioned methods have demonstrated precise RVs and FAZ segmentation in OCTA images. However, more complex tasks, such as AV segmentation in local regions, face challenges that often result in vessel misclassifications and discontinuities.

\subsection{Improvement of Segment Anything Model} 
\label{improve_sam}

SAM is a foundational vision model for general image segmentation, with the ability to segment diverse objects, parts, and visual structures in various scenarios \citep{kirillov2023segment}. SAM consists of three parts: an image encoder, a flexible prompt encoder, and a fast mask decoder. The image encoder utilizes a ViT pre-trained with the masked auto-encoder method; the prompt encoder processes points, bounding boxes, or coarse masks as prompts to support more local or zero-shot semantic segmentation; and the mask decoder generates segmentation predictions based on confidence scores. Although SAM has established an efficient data engine for model training, relatively few cases are collected for medical applications or other rare image scenarios. Therefore, some methods have been applied to SAM to improve its performance. 

SAMed applies the low-rank-based fine-tuning strategy to SAM on the labeled medical image segmentation datasets \citep{zhang2023customized}. SAM-Adapter integrates domain-specific information or visual prompts into the segmentation network using simple yet effective adapters to merge task-specific knowledge with the acquired general knowledge \citep{chen2023sam}. AutoSAM replaces the SAM's prompt encoder with a custom one to turn SAM into a fully automatic method \citep{shaharabany2023autosam}. \citet{chai2023ladder} propose to combine a complementary convolutional neural network (CNN) along with SAM for medical image segmentation. CWSAM freezes most of SAM’s parameters and incorporates lightweight adapters for fine-tuning, and a classwise mask decoder is designed to achieve semantic segmentation tasks \citep{pu2024classwise}. SAM-PARSER obtains the bases by matrix decomposition and fine-tuning the coefficients to reconstruct the parameter space tailored to the new scenario by an optimal linear combination of the bases \citep{peng2023sam}.  HQ-SAM designs a learnable high-quality output token, which is injected into SAM's mask decoder to predict the high-quality mask \citep{ke2023segment}.

These methods share the common characteristic of adding only a small number of trainable parameters when processing datasets in a new domain. It helps preserve the SAM's features of strong zero-shot capability and flexible extension.

\subsection{Artery-vein Segmentation}

Retinal arteries and veins are small blood vessels associated with various eye and systemic diseases \citep{cheung2015clinical}. Diseases impact arteries and veins differently. For instance, increased arteriolar tortuosity is linked to diabetic retinopathy, while widened venular caliber is associated with the risk of ischemic stroke. Hence, extracting and distinguishing retinal arteries and veins is crucial for their diagnostic significance \citep{seidelmann2016retinal, cheung2017retinal}.

Methods have been investigated for this problem of fundus and OCTA images. VTG-Net classifies retinal arteries and veins by a CNN-based framework incorporating vessel topology information \citep{mishra2021vtg}. \citet{hu2022multi} proposes a multi-scale interactive AV discrimination network that is able to achieve higher classification accuracy. AV-Net adopts a fully convolutional network with a transfer learning process to segment AV on enface OCT and OCTA images \citep{alam2020av}.

Due to the morphological similarities between retinal arteries and veins and the complexities introduced by factors such as age, gender, and disease conditions, previous methods often encounter segmentation disconnections or confusion in AV segmentation tasks.

%% file: sections/3_method.tex
\section{Method}

\begin{figure*}
    \centering
    \includegraphics[width=1\textwidth]{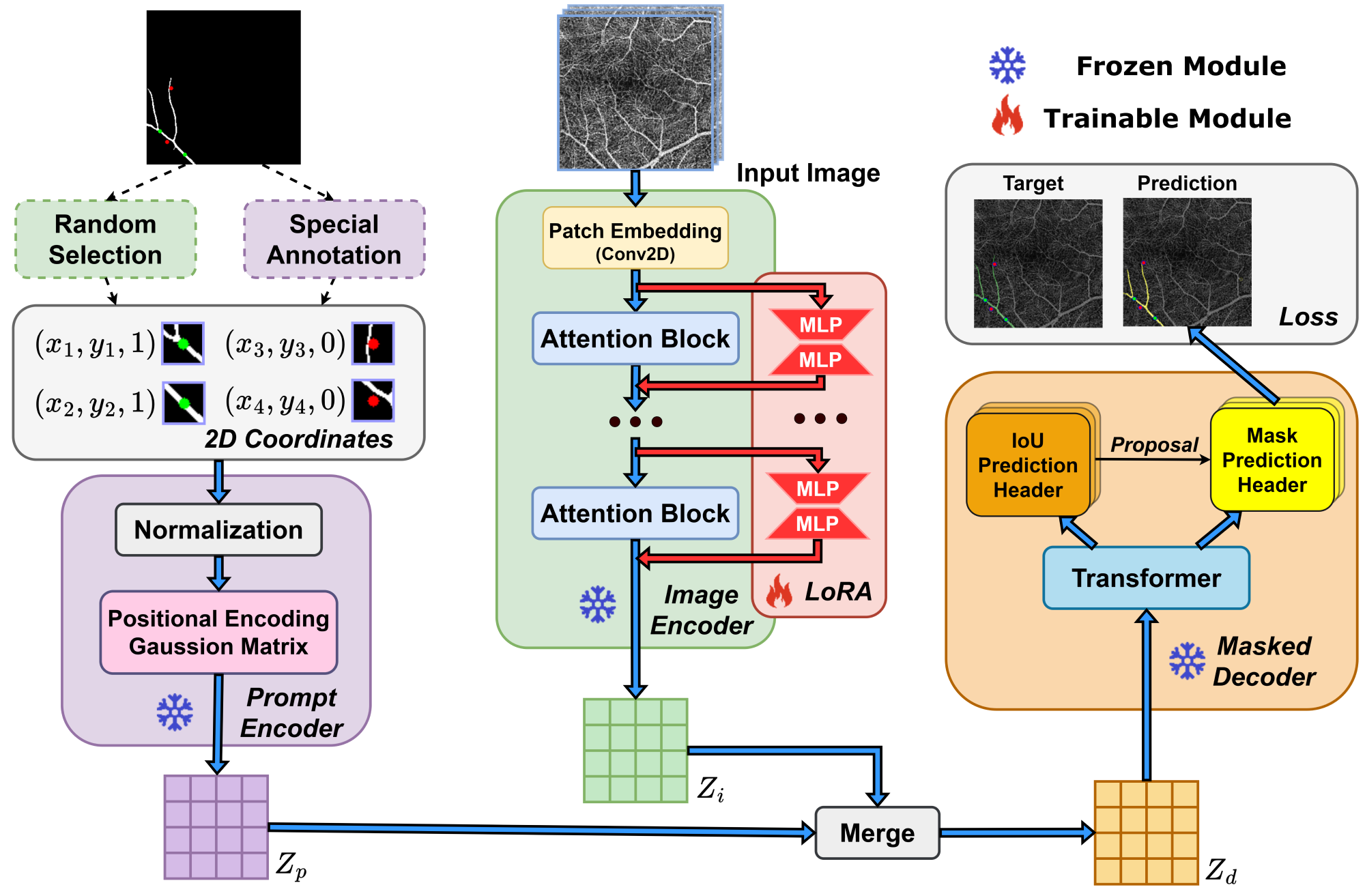}
    \caption{Schematic diagram illustrating the fine-tuning of SAM using OCTA samples.}
    \label{fig_sam_octa_architecture}
\end{figure*}

In this paper, we design a fine-tuning method based on SAM and a comprehensive generation strategy of prompt points for OCTA enface segmentation. The proposed method is named SAM-OCTA and is illustrated in Figure \ref{fig_sam_octa_architecture}. As an extension to the conference method, we have incorporated an additional prompt point generation strategy named special annotation. The segmentation tasks can be divided into two modes: global and local. In the global mode, all targets in the image should be segmented, whereas only the single object specified by the prompt points will be segmented in the local mode.

\subsection{Fine-tuning Image Encoder}

The ViT of the image encoder has three variants: ViT-b, ViT-l, and ViT-h, which can only process fixed-size inputs ($1024 \times 1024 \times 3$ and $224 \times 224 \times 3$). Scaling and padding operations are employed to support input images of different resolutions. In this study, the "ViT-h" is employed in experiments for its largest number of parameters if not otherwise specified.

\begin{figure}
    \centering
    \includegraphics[width=0.9\linewidth]{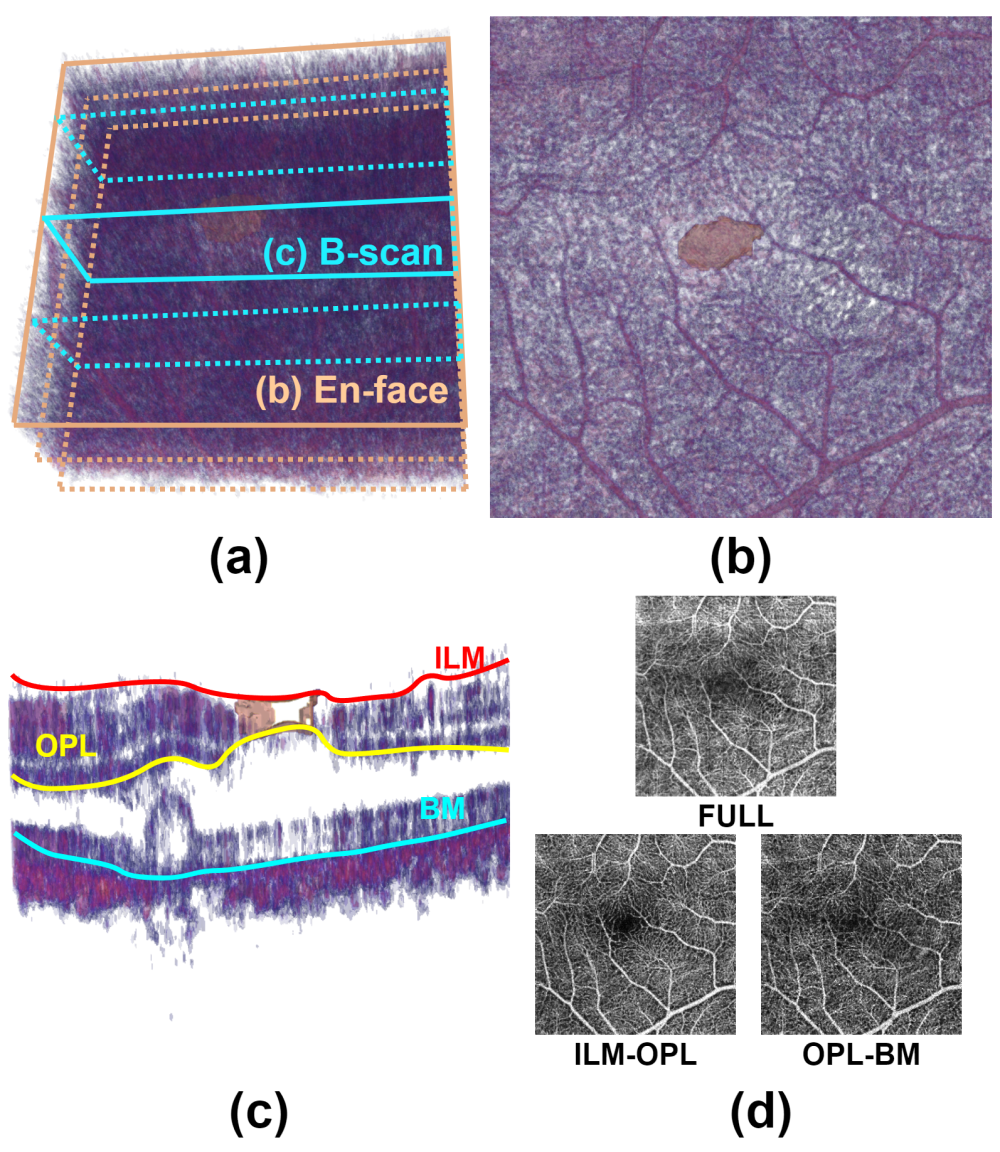}
    \caption{OCTA structural diagram. (a) Three-dimensional volume rendering. (b) En-face projection. (c) B-scan with annotated anatomic layers: internal limiting membrane (ILM), outer plexiform layer (OPL), and Bruch’s membrane (BM). (d) An en-face OCTA sample of the OCTA-500 dataset.}
    \label{fig_octa_data_structure}
\end{figure}

As shown in Figure \ref{fig_octa_data_structure}, OCTA data is inherently in 3D format. The enface OCTA images are 2D projections obtained by layer-wise segmentation based on anatomical structures. As SAM requires three-channel images as input, in this work, we stack en-face projection layers of OCTA images in different depths to adapt to this input format. The benefit of this approach is that it preserves the vascular structure information in the OCTA images while fully utilizing SAM's feature-extracting capabilities. 

Fine-tuning aims to retain SAM's powerful image understanding capabilities while enhancing its performance on the uncommon OCTA format. The approach used in this paper involves utilizing the low-rank adaptation (LoRA) technique \citep{hu2021lora}, which introduces additional linear network layers in each transformer block of SAM, similar in form to a ResNet block. During the training process, the image encoder's weights are frozen, and only the additional introduced parameters are updated. The fine-tuning process can be described as Algorithm \ref{algo-fine-tuning}:

\input{algorithms/sam_fine_tunning}

\subsection{Prompt Points Generation Strategy}

The prompt encoder receives two types of prompts as input: sparse prompts (points, boxes, text) and dense prompts (masks). In our work, we chose points as the prompt. Assuming there are $n$ points for each sample in the prompt point input, it can be represented as

$(x_1, y_1, 1), (x_2, y_2, 1), ..., (x_{n-1}, y_{n-1}, 0), (x_n, y_n, 0)$

\begin{figure}
    \centering
    \includegraphics[width=0.8\linewidth]{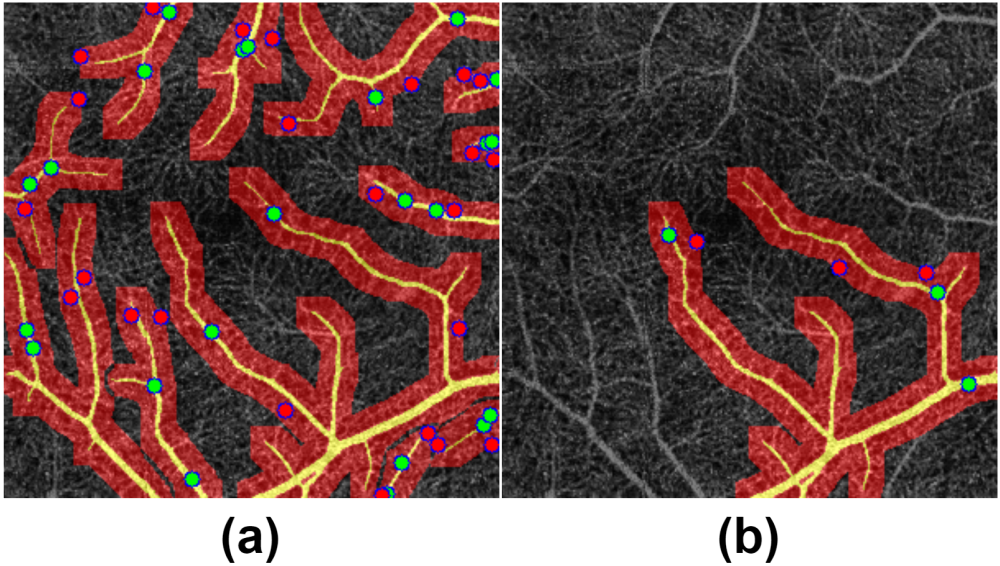}
    \caption{Two prompt points generation modes. (a) Global mode. (b) Local mode.}
    \label{fig_prompt_modes}
\end{figure}

The SAM defines 1 and 0 represent the positive and negative types of prompt points, respectively, corresponding to the target and background of the segmentation. The prompt encoder performs embedding on the coordinate input, and due to the pre-training, it can appropriately integrate with the information from the image encoder. The generation strategies have two types: global and local modes, as shown in Figure \ref{fig_prompt_modes}. Their difference lies in the target of segmentation. In the global mode, the generation process is applied to all connected components, while in the local mode, it is specific to a single connected component. The prompt generation strategies contain the following three types: random selection and special annotation.

\subsubsection{Random Selection}

\begin{figure*}
    \centering
    \includegraphics[width=1\linewidth]{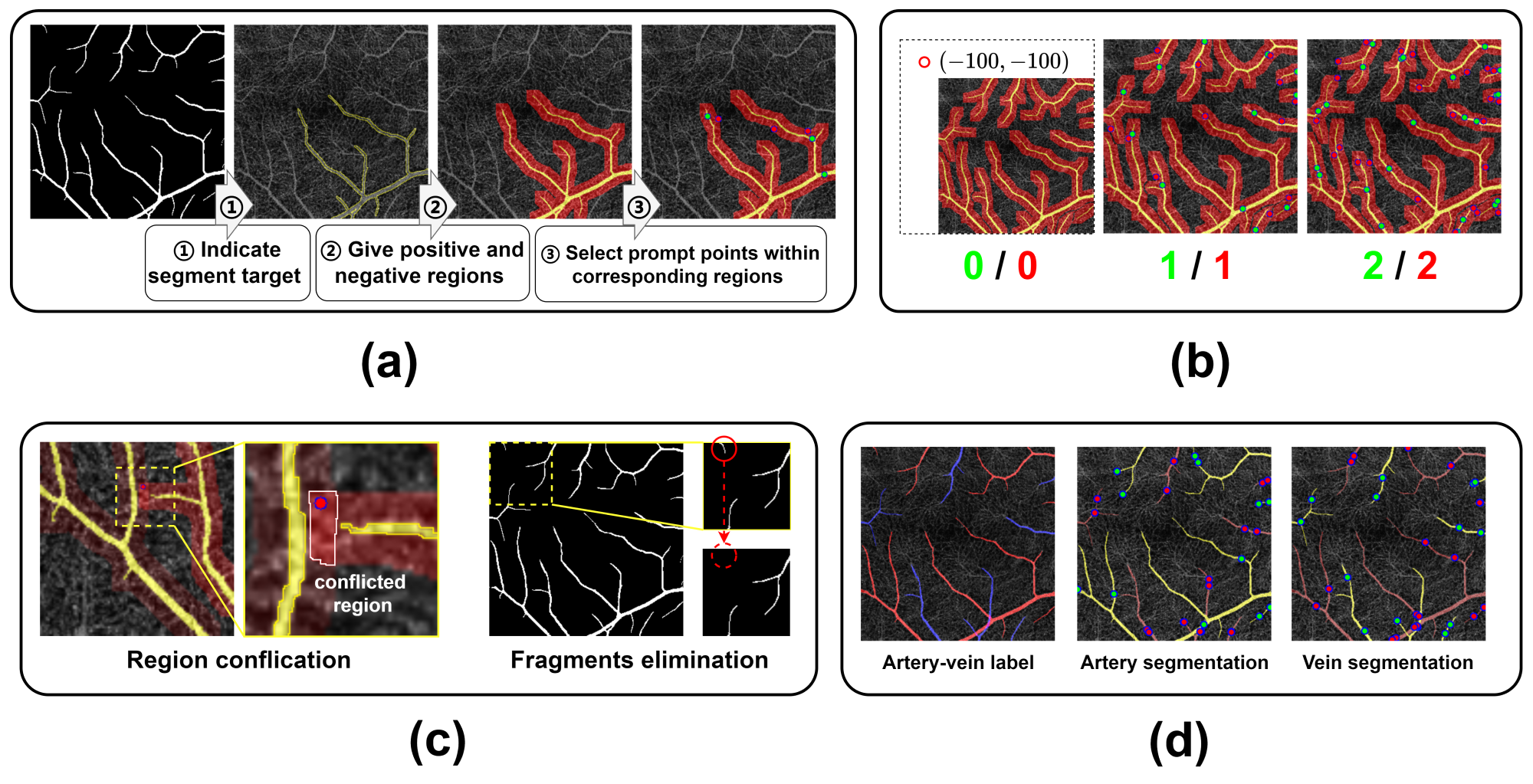}
    \caption{Schematic of random selection prompt method. (a) Positive and negative regions generation. (b) Prompt points selection of increasing quantity. (c) Handling of irregular cases. (d) Special strategy for artery-vein segmentation in global mode. }
    \label{fig_random_selection}
\end{figure*}

The overall process of selecting random prompt points consists of three steps as Figure \ref{fig_random_selection}(a). First, regions of each connected component are calculated for positive prompt points selection. Then, the depth-first-search algorithm is adopted to split out the negative regions along the connected components. The search max distance range is $d_{max}$ in pixels. Finally, the positive and negative points are randomly generated from corresponding regions. The purpose of randomly selecting points is to enable the model's generalization ability and construct comparable conditions for experiments.  The position of prompt point generation can be controlled by adjusting the range of proposal regions. In general, positive and negative points are selected equally for each connected component and its surrounding negative region as Figure \ref{fig_random_selection}(b). Specifically, when prompt points are not used, constant values are employed for filling (e.g. (-100, -100)). Figure \ref{fig_random_selection}(c) illustrates two cases of handling irregular situations. When a coordinate is simultaneously in the negative region of multiple connected components, it belongs to the nearest one. If the distances are equal, this coordinate is banned for prompt generation to avoid ambiguity. Some excessively small connected components might be due to annotation errors and are thus discarded (pixel area less than 20).

Furthermore, an extra generation strategy is made for AV segmentation tasks based on the corresponding nature in Figure \ref{fig_random_selection}(d). In this kind of task, we need to explore whether negative prompt points assist in distinguishing between arteries and veins. If the segmentation targets are arteries, the positive points are generated on arteries, and the negative points are selected from veins instead of the original negative region. When the segmentation target changes to veins, vice versa.

\subsubsection{Special Annotation}

\begin{figure}
    \centering
    \includegraphics[width=1\linewidth]{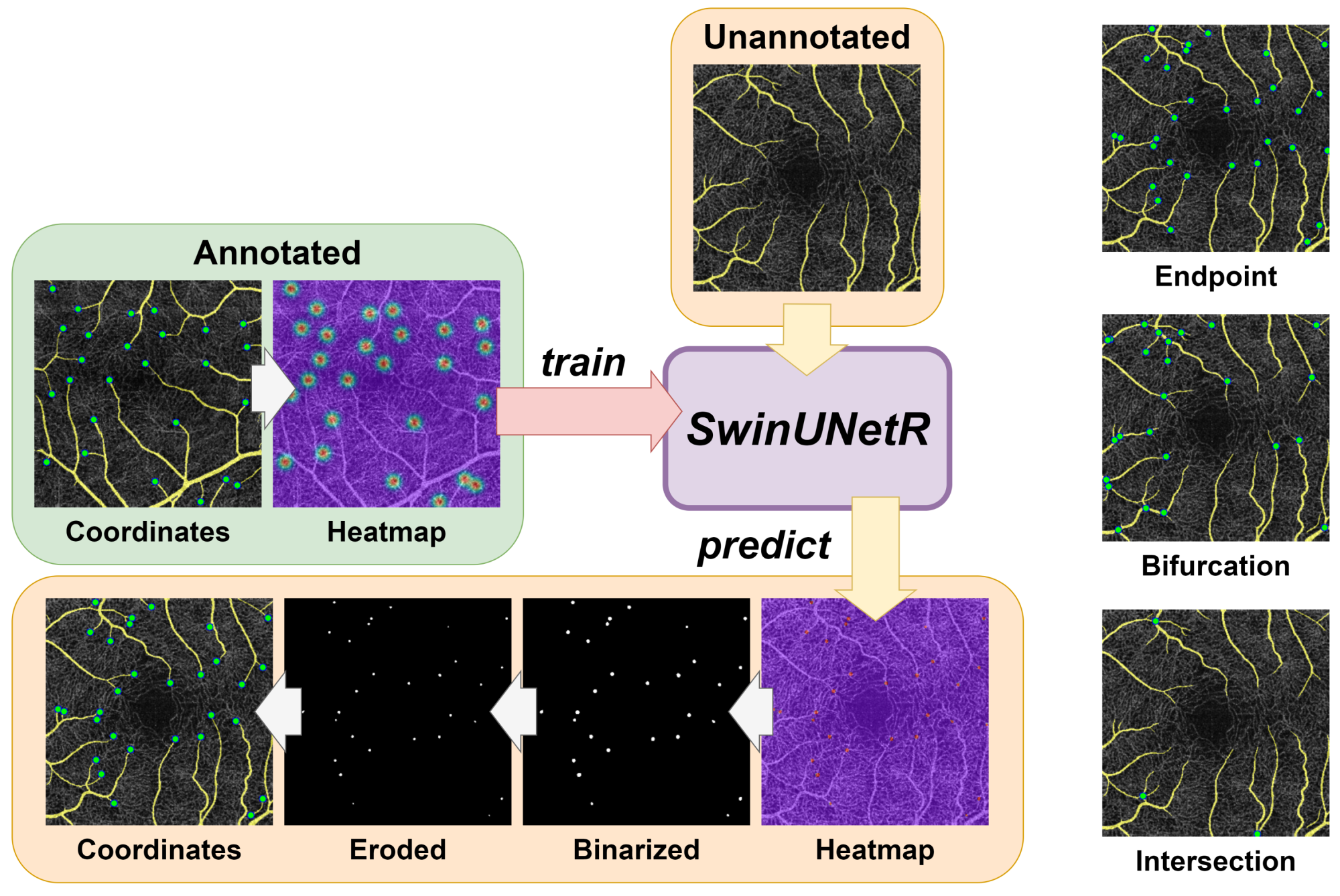}
    \caption{Sparse annotation method for special points based on heatmaps.}
    \label{fig_special_annotation}
\end{figure}

Endpoints, bifurcation, and intersection of retinal vascular trees play an important role in the diagnosis of numerous diseases or vessel occlusion \citep{hervella2020deep, wang2023automatic, xu2022mcg}. In this work, We collectively refer to these points as special points. It is worth exploring whether using these special points with medical significance as prompt points is helpful for vessel segmentation. Due to the dataset used in this paper lacking the relevant annotations, we adopt a sparse annotation approach to acquire the prompt points shown in Figure \ref{fig_special_annotation}.

Sparse annotation involves annotating only a small subset of samples in a dataset, minimizing the number of labeled instances to reduce annotation costs. We first manually annotated the positions of special points for $\frac{1}{10}$ of the samples in the dataset as the training set. The point coordinates are converted into Gaussian heatmaps, transforming the coordinate regression into segmentation to reduce prediction errors \citep{newell2016stacked}. The segmentation model adopted is SwinUNETR, which is implemented by MONAI \citep{hatamizadeh2021swin, cardoso2022monai}. SwinUNETR's encoder extracts features in multiple resolutions with shifted windows for computing self-attention and is connected to an FCNN-based decoder at each resolution via skip connections. The model has been proven to possess excellent feature extraction capabilities in wide medical image-related tasks. Apply the trained model to predict unlabeled samples to obtain the coordinates of special points after thresholding, erosion, and centroids calculating connected components. Finally, only very few annotation errors need to be inspected and corrected manually.

%% file: algorithms/sam_fine_tunning.tex
\begin{algorithm}
\caption{Fine-tunning procedure}
\label{algo-fine-tuning}
\begin{algorithmic}[1]  
\Require {
    \Statex $E_i$: the image encoder of SAM
    \Statex $E_p$: the prompt encoder of SAM
    \Statex $D$: the mask decoder of SAM
    \Statex $r$: rank of LoRA
    \Statex $x_i$: input image
    \Statex $x_p$: prompt points coordinates
    \Statex $y_{label}$: groud-truth of segmentation
}
\State Freeze the parameters of SAM's modules.
\State Pre-process the input image $x_i$ as $x$.
\For{$ab$ in $E_i.attention_blocks$}
\State $d = ab.qkv.dim$ // number of dimensions of qkv matrix
\State $l_{qa} = l_{va} = Linear(d, r)$
\State $l_{qb} = l_{vb} = Linear(r, d)$
\State $y_1 = ab.qkv(x)$
\State $y_q = l_{qb}(l_{qb}(x))$
\State $y_v = l_{vb}(l_{vb}(x))$
\State $x = y_1 + y_q + y_v$ // reassign x for next iteration
\EndFor
\State $z_i = x$ // latent feature of input image
\State $z_p = E_p(x_p)$ //convert prompt points coordinates to embeddings
\State Merge $z_i$ and $z_p$ as $z_d$.
\State $y_{mask} = D(z_d)$
\State Post-process the predicted mask $y_{mask}$.
\State Calculate the loss function based on $y_{label}$ and $y_{mask}$, and update the parameters of the LoRA module.
\end{algorithmic}
\end{algorithm}

%% file: sections/4_experiments.tex
\section{Experiments}

\subsection{Datasets and Preprocessing}

The dataset used in the experiments of this paper is OCTA-500, which is currently the most comprehensively annotated with enriched samples publicly available OCTA dataset \citep{li2020ipn}. It contains 500 samples, classified based on the field of view (FoV) with two scales: $3mm \times 3mm$ (3M) and $6mm \times 6mm$ (6M). The corresponding image resolutions are $304 \times 304$ and $400 \times 400$, with 200 and 300 samples respectively. Compared to the 3M images, the vessels in the 6M images are more complicated, and the FAZ is smaller. Therefore, the segmentation tasks of 6M are more challenging compared to 3M. The OCTA-500 dataset annotated labels on en-face images include FAZ, RV, capillary, artery, and vein. We have conducted experiments for the segmentation task of each label type.

The adopted data augmentation tool is Albumentations, which offers spatial- and pixel-level image transforms \citep{info11020125}. The data augmentation strategies for OCTA images include vertical and horizontal flipping, brightness adjustment, contrast-limited adaptive histogram equalization, and blur with a probability of 0.3. To better compare with other related works, we adopted the data partition strategy as \citet{li2020ipn}, dividing the dataset into training, test, and validation sets. The training set is utilized for adjusting model parameters, the test set evaluates fine-tuned performance, and the validation set is employed for early stopping to prevent overfitting. The partitioning details are summarized in Table \ref{Table_Partition}. In the evaluation of the test set and validation set, all connected components are included in the metrics calculation.

\begin{table}[t]
\centering
\caption{Partition of OCTA-500 dataset(by sample ID).}
\label{Table_Partition}
\begin{tabular}{cccc}
\toprule
FoV & Training Set & Validation Set & Test Set \\
\midrule
3M & 10301-10440 & 10441-10450 & 10451-10500 \\
6M & 10001-10180 & 10181-10200 & 10200-10300 \\
\bottomrule
\end{tabular}
\end{table}

\subsection{Experimental Settings}

SAM is deployed on A100 graphic cards with 80 GB of memory. The number of training epochs is 50. The optimizer used is AdamW, and the learning rate is set using a warm-up strategy, which can be expressed as: 

\begin{equation}
lr =
\begin{cases}
    \frac{t}{10^4}, t \leq 10 \\
    \max(\frac{1}{10^5}, \frac{1}{10^{4 * 0.98 * (t-10)}}), t > 10
\end{cases}
\end{equation}

    where,

    $lr$ → {\itshape learning rate},
    
    $t$ → {\itshape trained epochs}.

The loss functions used for fine-tuning vary depending on the segmentation tasks. For FAZ and capillary, the Dice loss is employed. However, for RV, artery, and vein, the clDice loss is utilized \citep{shit2021cldice}. The clDice loss can leverage centerline topological property to assist in preserving the connectivity of tubular segmentations such as vessels and roads. Min- and max-pooling layers are used to iteratively extract the centerline of the predicted image and the label, similar to the erosion operation in morphology. Then, based on the predicted values, labels, and the extracted centerline, the clDice loss can be calculated. These two loss functions can be represented as: 

\begin{equation}
L_{Dice} = 1 - \frac{2 * |\hat{Y} \cap Y|}{|\hat{Y}| + |Y|},
\end{equation}

\begin{equation}
T{prec}(\hat{Y_s}, Y) = \frac{|\hat{Y_s} \odot Y| + \epsilon}{|\hat{Y_s}| + \epsilon},
\end{equation}

\begin{equation}
T{sens}(Y_s, \hat{Y}) = \frac{|Y_s \odot \hat{Y}| + \epsilon}{|Y_s| + \epsilon},
\end{equation}

\begin{equation}
L_{clDice} = 1 - 2 * \frac{T{prec}(\hat{Y_s}, Y) * T{sens}(Y_s, \hat{Y})}{T{prec}(\hat{Y_s}, Y) + T{sens}(Y_s, \hat{Y})},
\end{equation}

    where

    $Y$ → {\itshape the ground-truth},
    
    $\hat{Y}$ → {\itshape the predicted value},

    $Y_s, \hat{Y_s}$ → {\itshape soft{-}skeleton($Y$, $\hat{Y}$)},

    $\epsilon$ → {\itshape $10^{-6}$ }.

Experiments conducted in previous studies have shown that using only the clDice loss can lead to the failure of the model training. In practical applications, it is necessary to use a weighted combination of clDice loss and Dice loss for effective results:

\begin{equation}
L_{clDice} = 0.8 * L_{Dice} + 0.2 * L_{clDice}^\prime.
\end{equation}

We conducted experiments about the three prompt point generation strategies. The segmentation results using metrics Dice, Jaccard, and Hausdorff Distance (HD), which are calculated as follows:

\begin{equation}
    Dice(\hat{Y}, Y) = \frac{2 |\hat{Y} \cap Y|}{|\hat{Y}| + |Y|},
\end{equation}

\begin{equation}
    Jaccard(\hat{Y}, Y) = \frac{|\hat{Y} \cap Y|}{|\hat{Y} \cup Y|},
\end{equation}

\begin{equation}
    HD(\hat{Y}, Y) = \max(h(\hat{Y}, Y), h(Y, \hat{Y})),
\end{equation}

\begin{equation}
    h(\hat{Y}, Y) = \max_{\hat{y} \in \hat{Y}} \min_{y \in Y} ||\hat{y} - y||.
\end{equation}

\subsection{Results}

\subsubsection{Random Selection} \label{section_results_random_selection}

We initially explore the effect of the number, types, and positions of prompt points on segmentation results using the "random selection" generation strategy. 0 to 5 prompt points are randomly generated for each connected component in global segmentation tasks, and 1 to 8 are generated in local segmentation. In local mode, at least one prompt point should be retained to indicate the connected component of the local region; otherwise, segmentation ambiguity may occur. In both modes, the $d_{max}$ values 9. 

\begin{figure*}
    \centering
    \includegraphics[width=1\textwidth]{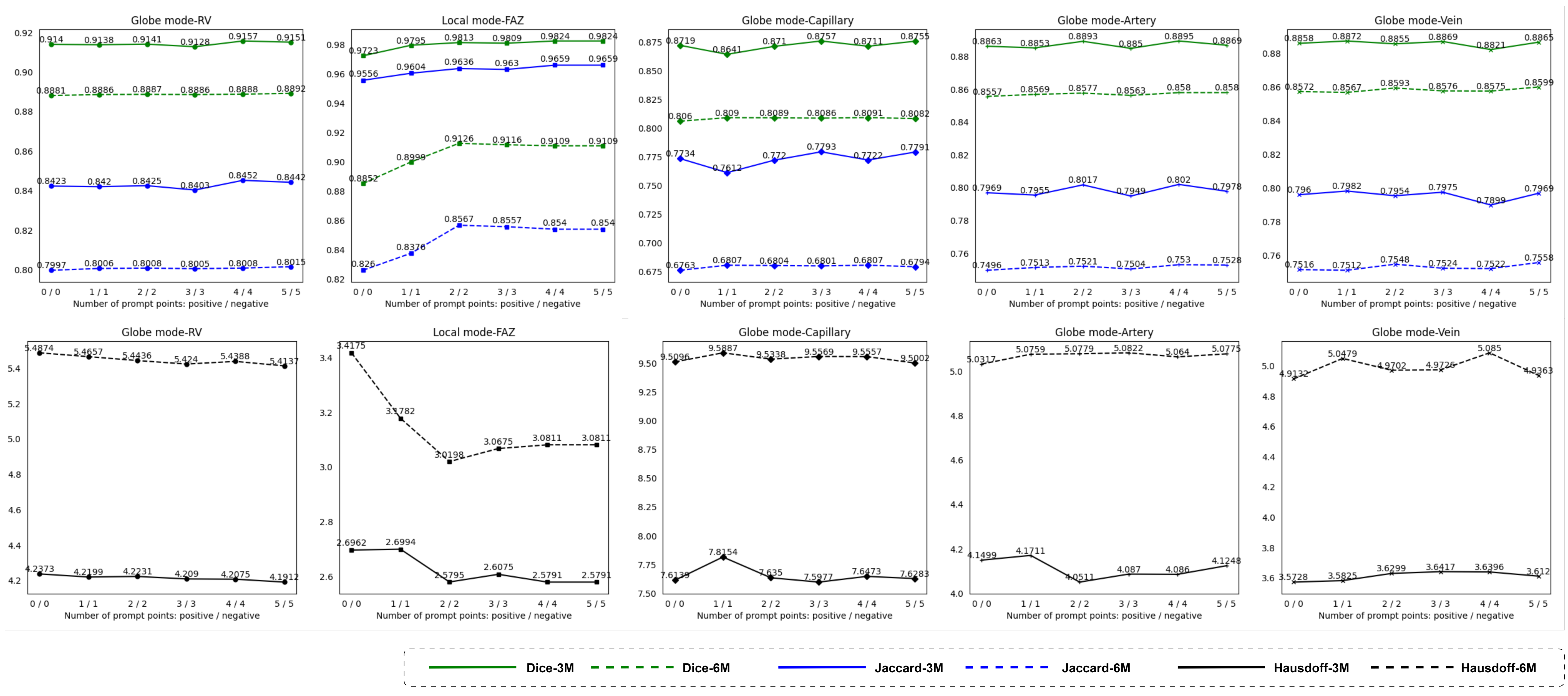}
    \caption{Random selection results under the increasing number of prompt points (global mode).}
    \label{fig_result_glocal_metrics}
\end{figure*}

\begin{figure}
    \centering
    \includegraphics[width=1\linewidth]{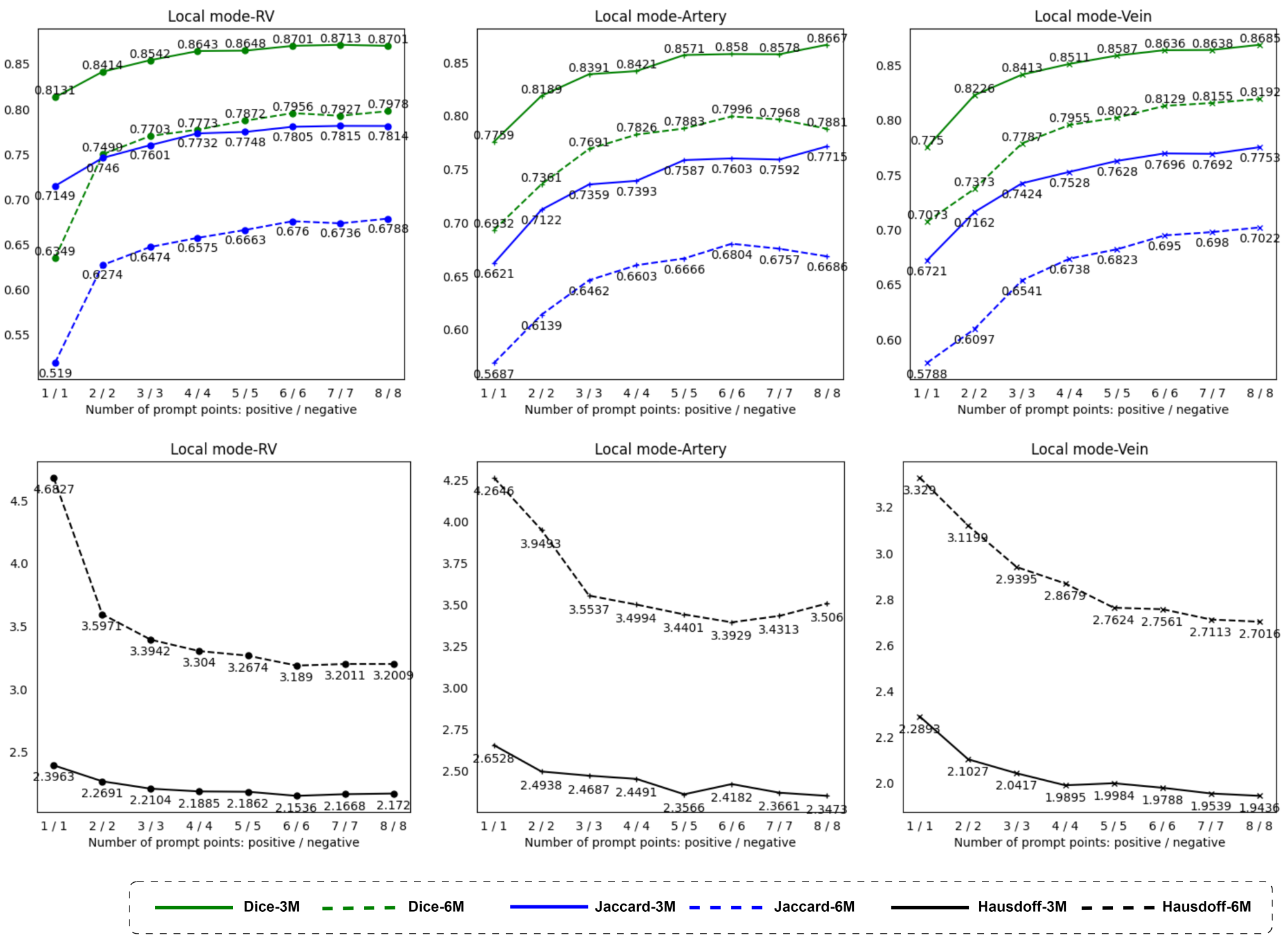}
    \caption{Random selection results under the increasing number of prompt points (local mode).}
    \label{fig_result_local_metrics}
\end{figure}

\begin{figure*}
    \centering
    \includegraphics[width=0.95\textwidth]{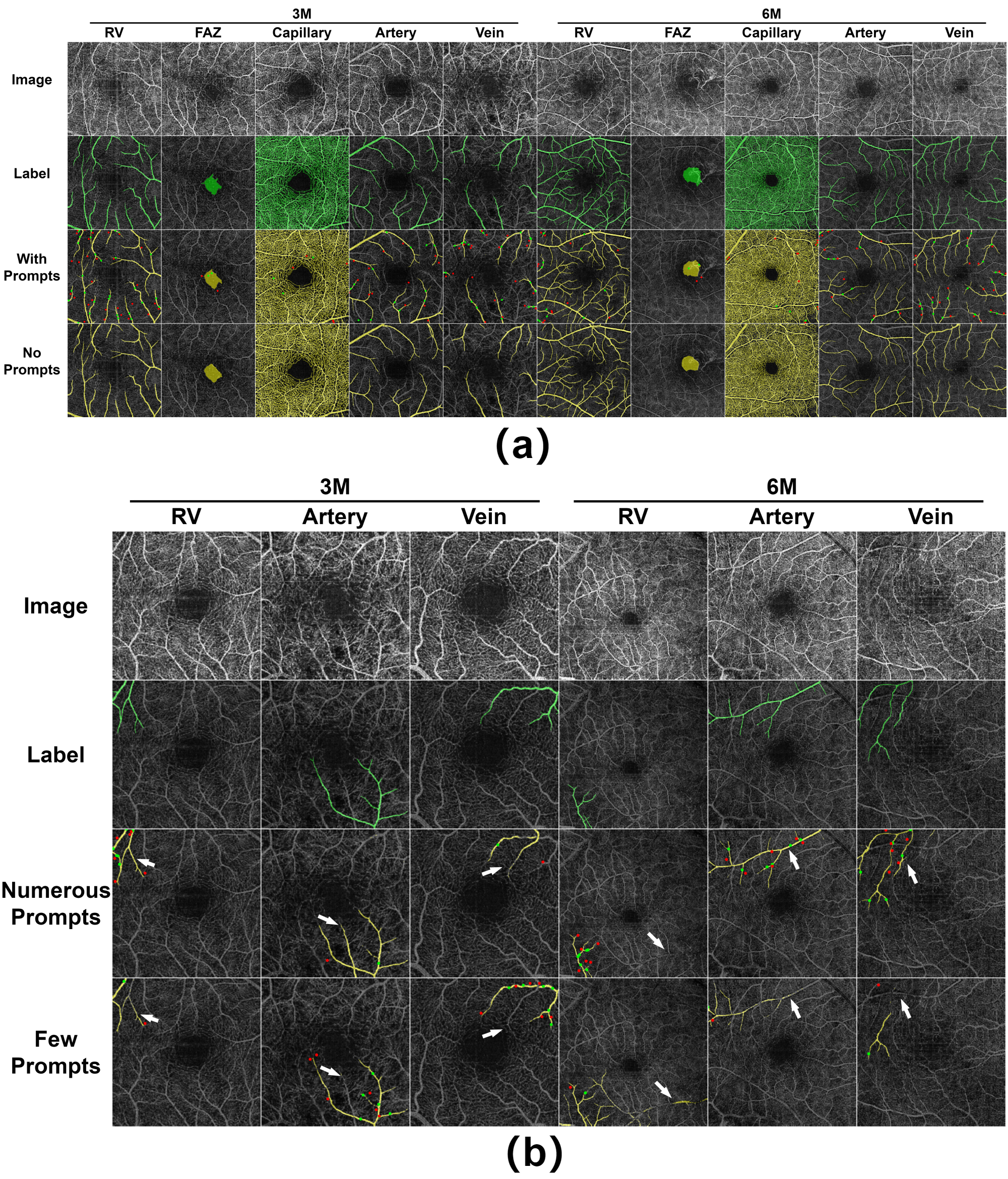}
    \caption{Segmentation examples using the random selection strategy. (a) Global mode. (b) Local mode. White arrows indicate the difference in segmentation outcomes for the number of prompt points.}
    \label{fig_result_seg}
\end{figure*}

The experimental results under global mode and local mode are summarized in Figure \ref{fig_result_glocal_metrics} and \ref{fig_result_local_metrics}. It can be observed that, except for FAZ, the prompt points have little impact on the segmentation metrics in global mode. In fact, FAZ is unique because its target contains only a single connected component. Therefore, the segmentation task of FAZ in both the global mode and the local mode is the same. The segmentation performance of each task in local mode improves with an increase in the number of prompt points. This indicates that prompt points play a significant role in the local mode. The segmentation performance improves rapidly with increased points when prompt points are few. When the number of positive and negative prompt points reaches six or more each, most of the task metrics tend to converge or even decrease. Figures \ref{fig_result_seg} displays segmentation examples of tasks in two different modes. For global segmentation, it can be observed that even without prompt points, the fine-tuned SAM performs as well as in cases with prompt points. In local mode, the target vessel, such as AV segmentation with 6M FoV, may not be accurately identified and segmented with fewer prompt points. Additionally, some non-target blood vessels may be incorrectly detected and can be observed in 3M AV segmentation. The increased number of prompt points clearly delineates segmentation regions, avoiding ambiguity in local segmentation. This is a principal factor contributing to the numerical enhancement of segmentation metrics.

We also adjust the generation strategy for negative points in the global mode as Figure \ref{fig_random_selection}(d) to explore their potential roles. This strategy is implemented in AV segmentation. For artery segmentation, three positive and negative points are generated on each artery and vein correspondingly, and vice versa in vein segmentation. The experiment is to test whether negative points have a segmentation-suppressing effect on non-target vessels, and the results are shown in Table \ref{table-nontarget_result}. With the equal numbers of prompt points, segmentation performance slightly decreases under this strategy compared to cue points selected from negative regions. 

\input{tables/nontarget_result}

We further explored the contribution of positive and negative prompt points to segmentation and conducted ablation experiments with only positive or negative prompt points in the local mode. The experiment was conducted under the condition of prompt points number with 6 for each type, and the results are summarized in Table \ref{table-prompt_type_result}. The segmentation metrics are highest when both positive and negative prompt points are used, except for RV segmentation with 3M FoV. Actually, with only positive points applied, the segmentation performance is very close to using both types of prompt points. When only negative points are used, the segmentation performance noticeably decreases. This is even inferior to the case of using both positive and negative points with the same total quantity.

\input{tables/prompt_type_result}

\subsubsection{Special Annotation}

\begin{figure}
    \centering
    \includegraphics[width=1\linewidth]{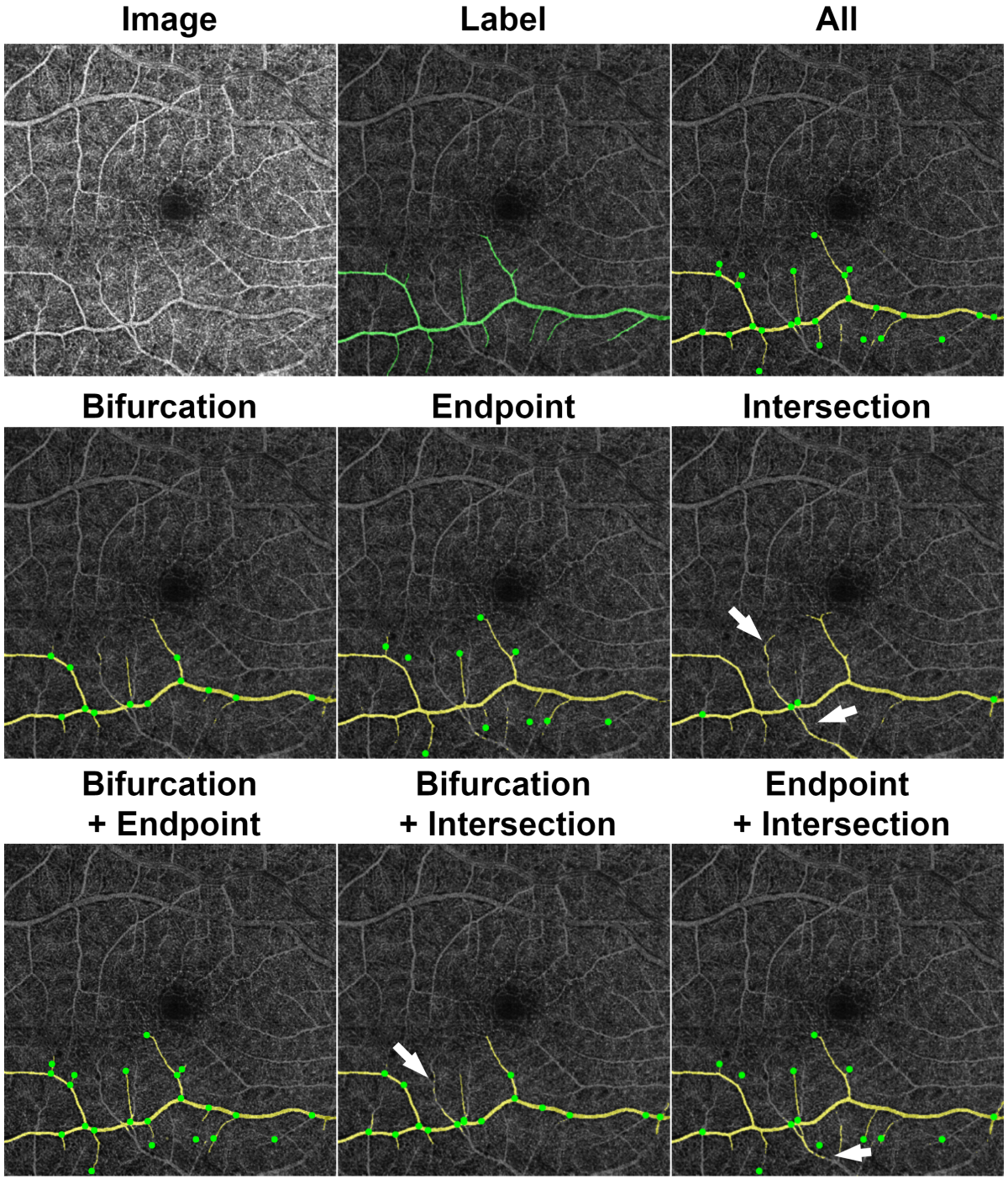}
    \caption{Illustration of local arterial segmentation using special points and their combinations as positive prompt points. White arrows indicate some erroneously segmented vessel branches.}
    \label{fig_result_special_points}
\end{figure}

The generation of special points is fixed, and experiments are conducted using combinations of special points from different positions. All special points adopted as prompt points enable effective local vessel segmentation. From the results of Section \ref{section_results_random_selection}, it's evident that prompt points provide minimal enhancement to segmentation performance under the global mode, and generating inappropriate points may lead to degradation. Furthermore, bifurcations, endpoints, and intersections are all located above the vessels. Therefore, this experiment is conducted in local mode, with the prompt points type being positive. Combining different types of special points, such as endpoints and bifurcations, may improve segmentation performance in most cases. However, selecting only bifurcations can yield better performance than selecting both bifurcations and intersections, except for cases of vein segmentation with 6M FoV. From the artery segmentation example in Figure \ref{fig_result_special_points}, it can be observed that intersections can lead the model to misclassify the type of artery and vein. In the sub-figures containing intersections, the model segments some venous branches as arteries. From Table \ref{table-special_points_result}, it can be observed that combining all special points generally leads to the best segmentation metrics, consistent with the conclusion of local segmentation in random selection experiments.

Contrary to the random selection generation strategy, the number of special points corresponding to each vessel varies. Long vessels typically contain more prompt points, and the number of endpoints is also significantly greater than that of bifurcations and intersections. The reason is that the appearance of bifurcation points also increases the endpoints of blood vessels, according to the topological relationship of the graph theory, when the vessels are intact. Furthermore, the intersections of the arteries and veins are rarer compared to bifurcation points. For instance, in artery tasks with 3M FoV, each target averages 1.9 bifurcations, 2.2 endpoints, and 1.1 intersections. Although adopting intersections performs worse than adopting bifurcations and endpoints in segmentation, it requires fewer prompt points. In contrast to Figure \ref{fig_result_local_metrics}, selecting special points typically yields better performance compared to random generation if the same average quantity of prompt points is provided. 

\input{tables/special_points_result}

\subsection{Comparative Results}

To provide a more objective assessment of the performance of our SAM-OCTA method, we compared it with several well-known methods for OCTA segmentation tasks. RV and FAZ are common segmentation tasks in previous studies. Therefore, we will summarize the comparative results in Table \ref{Table_Result_Comparison}, and the best metrics have been selected as our final results. The experimental data from previous methods are referenced from \cite{hu2022joint}. Our method's comprehensive performance reaches the state-of-the-art level.

\begin{figure}
    \centering
    \includegraphics[width=1\linewidth]{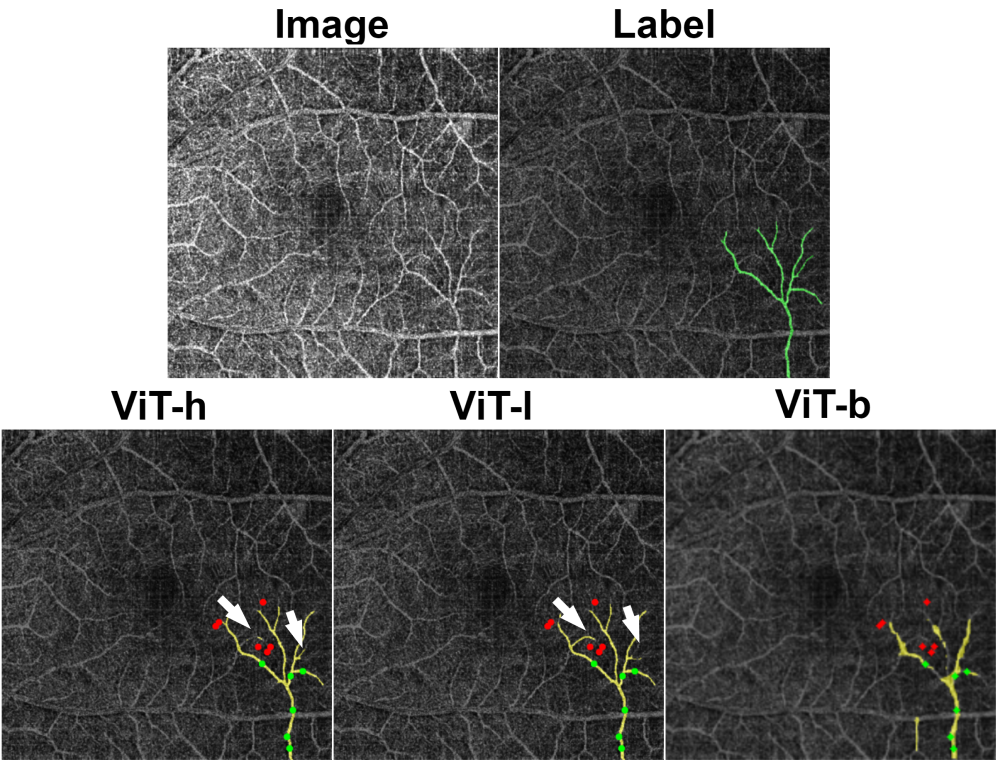}
    \caption{Illustration of local vein segmentation using ViT of different scales as image encoders. White arrows indicate some erroneously segmented vessel branches.}
    \label{fig_result_model_scales}
\end{figure}

Fine-tuning is also conducted for these segmentation tasks in smaller-scale SAMs. This provides a more practical reference for the effectiveness of SAM-OCTA. The experimental results are summarized in Tables \ref{table-scales_result} at last page. The strategy used for generating prompt points is random selection. Due to the minimal impact of prompt points in global mode, the best results under different numbers of prompt points are adopted. In the local mode, six positive and six negative points are generated in each case. The ViT-h, having the largest number of parameters, exhibits the best performance in all tasks except for the 6M RV and 3M artery segmentation in local mode. The ViT-l's efficacy is remarkably similar to that of the ViT-h, particularly in tasks related to RV, artery, and vein. However, the segmentation performance of ViT-b is noticeably inferior. Taking the example of the 6M arterial local segmentation shown in Figure \ref{fig_result_model_scales}, ViT-h and ViT-l extract the overall shape of the vessels in alignment with the annotations. For the subtle vessel branch segmentation, ViT-h is more accurate than ViT-l, and ViT-b almost loses the ability to process the branching details. The vessels segmented by the ViT-b model only vaguely indicate the stem. The thickness of the vessels is largely inaccurate, and the segmented structures are highly fragmented. A larger model scale has a better segmentation effect, and this also leads to more memory usage and a longer fine-tuning duration. The size of ViT-l is three-quarters that of ViT-h, and the time required for fine-tuning is only half. In practical use, if the memory allows, it is best to use the ViT-h model for segmentation to achieve the best segmentation effect. Vit-l can also perform well in practical use as a lightweight option. Although ViT-b is the smallest and requires less time for fine-tuning, it is strongly not recommended for use due to its significantly lower resolution and effectiveness compared to the former two. It is nearly impossible to correctly reflect the information of biomarkers using ViT-b, which could lead to serious misjudgments in medical diagnosis.

\input{tables/result_comparison}

\input{tables/scales_result}

%% file: tables/nontarget_result.tex
\begin{table}[h]
\caption{Local vessel segmentation results with combinations of special points as prompt points (underscores indicate the optimal value).}
\setlength{\extrarowheight}{-2pt}
\label{table-nontarget_result}
\begin{tabular*}{\tblwidth}{@{}LLLLLL@{}}
\toprule
\multirow{2}{*}{Label} & \multirow{2}{*}{FoV} & \multirow{2}{*}{Dice ↑} & \multirow{2}{*}{Jaccard ↑} & \multirow{2}{*}{HD ↓} & Non-target\\
&&&&& Vessel \\
\midrule
\multirow{4}{*}{Artery} & \multirow{2}{*}{3M} & 0.8654 & 0.7646 & 4.2727 & \checkmark \\
& & \underline{0.8850} & \underline{0.7949} & \underline{4.0860} & \XSolidBrush \\
\cmidrule(lr){2-6}
& \multirow{2}{*}{6M} & 0.8439 & 0.7319 & 5.2216 & \checkmark \\
& & \underline{0.8563} & \underline{0.7504} & \underline{5.0822} & \XSolidBrush \\
\cmidrule(lr){1-6}
\multirow{4}{*}{Vein} & \multirow{2}{*}{3M} & 0.8620 & 0.7583 & 3.8494 & \checkmark \\
& & \underline{0.8869} & \underline{0.7975} & \underline{3.6417} & \XSolidBrush \\
\cmidrule(lr){2-6}
& \multirow{2}{*}{6M} & 0.8482 & 0.7381 & 5.1123 & \checkmark \\
& & \underline{0.8576} & \underline{0.7524} & \underline{4.9726} & \XSolidBrush \\

\bottomrule
\end{tabular*}
\end{table}

%% file: tables/prompt_type_result.tex
\begin{table}[h]
\caption{Local vessel segmentation results of single type prompt points.}
\setlength{\extrarowheight}{-2pt}
\label{table-prompt_type_result}
\begin{tabular*}{\tblwidth}{@{}LLLLLLL@{}}
\toprule
\multirow{2}{*}{Label} & \multirow{2}{*}{FoV} & \multicolumn{2}{L}{Prompt Points} & \multirow{2}{*}{Dice ↑} & \multirow{2}{*}{Jaccard ↑} & \multirow{2}{*}{HD ↓} \\ 
\cmidrule(lr){3-4}
& & Positive   & Negative & & & \\
\midrule
\multirow{8}{*}{RV} & \multirow{4}{*}{3M} & 6 & 6 & 0.8701 & 0.7805 & 2.1536 \\
& & \textbf{6} & \textbf{0} & \textbf{0.8725} & \textbf{0.7835} & \textbf{2.1345} \\
& & \textbf{0} & \textbf{6} & \textbf{0.8440} & \textbf{0.7493} & \textbf{2.2986} \\
& & 3 & 3 & 0.8542 & 0.7601 & 2.2104 \\
\cmidrule(lr){2-7}
& \multirow{4}{*}{6M} & 6 & 6 & 0.7956 & 0.6760 & 3.1890 \\
& & \textbf{6} & \textbf{0} & \textbf{0.7868} & \textbf{0.6662} & \textbf{3.2551} \\
& & \textbf{0} & \textbf{6} & \textbf{0.7545} & \textbf{0.6338} & \textbf{3.5532} \\
& & 3 & 3 & 0.7703 & 0.6474 & 3.3942 \\
\cmidrule(lr){1-7}
\multirow{8}{*}{Artery} & \multirow{4}{*}{3M} & 6 & 6 & 0.8580 & 0.7603 & 2.4182 \\
& & \textbf{6} & \textbf{0} & \textbf{0.8494} & \textbf{0.7493} & \textbf{2.3921} \\
& & \textbf{0} & \textbf{6} & \textbf{0.8116} & \textbf{0.7041} & \textbf{2.5738} \\
& & 3 & 3 & 0.8391 & 0.7359 & 2.4687 \\
\cmidrule(lr){2-7}
& \multirow{4}{*}{6M} & 6 & 6 & 0.7996 & 0.6804 & 3.3929 \\
& & \textbf{6} & \textbf{0} & \textbf{0.7950} & \textbf{0.6737} & \textbf{3.4283} \\
& & \textbf{0} & \textbf{6} & \textbf{0.7383} & \textbf{0.6196} & \textbf{3.8323} \\
& & 3 & 3 & 0.7691 & 0.6462 & 3.5537 \\
\cmidrule(lr){1-7}
\multirow{8}{*}{Vein} & \multirow{4}{*}{3M} & 6 & 6 & 0.8636 & 0.7696 & 1.9788 \\
& & 6 & 0 & \textbf{0.8604} & \textbf{0.7646} & \textbf{2.0000} \\
& & 0 & 6 & \textbf{0.8068} & \textbf{0.6989} & \textbf{2.2308} \\
& & 3 & 3 & 0.8413 & 0.7424 & 2.0417 \\
\cmidrule(lr){2-7}
& \multirow{4}{*}{6M} & 6 & 6 & 0.8129 & 0.6950 & 2.7561 \\
& & 6 & 0 & \textbf{0.8103} & \textbf{0.6919} & \textbf{2.7898} \\
& & 0 & 6 & \textbf{0.7584} & \textbf{0.6366} & \textbf{3.0852} \\
& & 3 & 3 & 0.7787 & 0.6541 & 2.9395 \\
\bottomrule
\end{tabular*}
\end{table}

%% file: tables/special_points_result.tex
\begin{table}[h]
\caption{Results of negative prompt points selected from non-target vessels in global AV segmentation (underscores indicate the optimal values).}
\setlength{\extrarowheight}{-3pt}
\label{table-special_points_result}
\begin{tabular*}{\tblwidth}{@{}LLLLLLLL@{}}
\toprule
Label & FoV & Dice ↑ & Jaccard ↑ & HD ↓ & Bif. & End. & Int. \\ 
\midrule
\multirow{6}{*}{RV} & \multirow{3}{*}{3M} 
  & 0.8451 & 0.7476 & 2.3104 & \checkmark & & \\
& & 0.8416 & 0.7431 & 2.3143 & & \checkmark & \\
& & \underline{0.8535} & \underline{0.7580} & \underline{2.2621} & \checkmark & \checkmark & \\
\cmidrule(lr){2-8}
 & \multirow{3}{*}{6M} & 0.7906 & 0.6671 & \underline{3.3175} & \checkmark & & \\
& & 0.7727 & 0.6508 & 3.4328 & & \checkmark & \\
& & \underline{0.7923} & \underline{0.6716} & 3.3238 & \checkmark & \checkmark & \\
\cmidrule(lr){1-8}
\multirow{14}{*}{Artery} & \multirow{7}{*}{3M} & 0.8244 & 0.7157 & \underline{2.4257} & \checkmark & & \\
& & 0.7979 & 0.6898 & 2.5896 & & \checkmark & \\
& & 0.7635 & 0.6500 & 2.6971 & & & \checkmark \\
& & 0.8284 & 0.7251 & 2.4341 & \checkmark & \checkmark & \\
& & 0.8096 & 0.6995 & 2.5043 & \checkmark & & \checkmark \\
& & 0.7961 & 0.6866 & 2.5940 & & \checkmark & \checkmark \\
& & \underline{0.8292} & \underline{0.7255} & 2.4437 & \checkmark & \checkmark & \checkmark \\
\cmidrule(lr){2-8}
 & \multirow{7}{*}{6M} & 0.7594 & 0.6289 & 3.6144 & \checkmark & & \\
& & 0.7411 & 0.6120 & 3.7210 & & \checkmark & \\
& & 0.7060 & 0.5740 & 3.7953 & & & \checkmark \\
& & \underline{0.7628} & \underline{0.6390} & 3.5991 & \checkmark & \checkmark & \\
& & 0.7354 & 0.6016 & 3.6889 & \checkmark & & \checkmark \\
& & 0.7305 & 0.6011 & 3.7257 & & \checkmark & \checkmark \\
& & 0.7627 & 0.6353 & \underline{3.5882} & \checkmark & \checkmark & \checkmark \\
\cmidrule(lr){1-8}
\multirow{14}{*}{Vein} & \multirow{7}{*}{3M} & 0.8119 & 0.7059 & 2.1862 & \checkmark & & \\
& & 0.8110 & 0.7090 & 2.1059 & & \checkmark & \\
& & 0.7444 & 0.6398 & 2.2989 & & & \checkmark \\
& & 0.8166 & 0.7127 & 2.0857 & \checkmark & \checkmark & \\
& & 0.8007 & 0.6917 & 2.1624 & \checkmark & & \checkmark \\
& & 0.8217 & 0.7189 & 2.0785 & & \checkmark & \checkmark \\
& & \underline{0.8281} & \underline{0.7238} & \underline{2.0782} & \checkmark & \checkmark & \checkmark \\
\cmidrule(lr){2-8}
 & \multirow{7}{*}{6M} & 0.7615 & 0.6302 & 3.0047 & \checkmark & & \\
& & 0.7400 & 0.6127 & 3.1202 & & \checkmark & \\
& & 0.7139 & 0.5849 & 3.2612 & & & \checkmark \\
& & 0.7710 & 0.6467 & 2.9423 & \checkmark & \checkmark & \\
& & 0.7658 & 0.6363 & 3.0208 & \checkmark & & \checkmark \\
& & 0.7608 & 0.6337 & 3.0953 & & \checkmark & \checkmark \\
& & \underline{0.7810} & \underline{0.6564} & \underline{2.9822} & \checkmark & \checkmark & \checkmark \\

\bottomrule
\end{tabular*}
\begin{tablenotes}
\item[*] Bif. : Bifurcation; End. : Endpoint; Int. : Intersection. 
\end{tablenotes}
\end{table}

%% file: tables/result_comparison.tex
\begin{table*}[h]
\centering
\caption{RV and FAZ segmentation results on OCTA-500 (underscores indicate the top two highest values).}
\label{Table_Result_Comparison}
\begin{tabular}{ccccccccc}
\toprule
Label & \multicolumn{4}{c}{RV} & \multicolumn{4}{c}{FAZ} \\
\midrule
\multirow{2}{*}{Method} & \multicolumn{2}{c}{3M} & \multicolumn{2}{c}{6M} & \multicolumn{2}{c}{3M} & \multicolumn{2}{c}{6M} \\
\cmidrule(lr){2-3} \cmidrule(lr){4-5} \cmidrule(lr){6-7} \cmidrule(lr){8-9}
& Dice ↑ & Jaccard ↑ & Dice ↑ & Jaccard ↑ & Dice ↑ & Jaccard ↑ & Dice ↑ & Jaccard ↑ \\
\midrule
    U-Net (2015) & 0.9068 & 0.8301 & 0.8876 & 0.7987 & 0.9747 & 0.9585 & 0.8770 & 0.8124 \\
      IPN (2020) & 0.9062 & 0.8325 & 0.8864 & 0.7973 & 0.9505 & 0.9091 & 0.8802 & 0.7980 \\
  IPN V2+ (2020) & \underline{0.9274} & \underline{0.8667} & \underline{0.8941} & \underline{0.8095} & 0.9755 & 0.9532 & \underline{0.9084} & 0.8423 \\
    FARGO (2021) & 0.9112 & 0.8374 & 0.8798 & 0.7864 & 0.9785 & 0.9587 & 0.8930 & 0.8355 \\
Joint-Seg (2022) & 0.9113 & 0.8378 & \underline{0.8972} & \underline{0.8117} & \underline{0.9843} & \underline{0.9693} & 0.9051 & \underline{0.8424} \\
\midrule
SAM-OCTA (ours) & \underline{\textbf{0.9157}} & \underline{\textbf{0.8452}} & 0.8898 & 0.8015 & \underline{\textbf{0.9824}} & \underline{\textbf{0.9659}} & \underline{\textbf{0.9126}} & \underline{\textbf{0.8547}} \\
\bottomrule
\end{tabular}
\end{table*}

%% file: tables/scales_result.tex
\begin{table*}[h]
\caption{Segmentation results of fine-tuned SAM with different model scales (underscores indicate the optimal values).}
\setlength{\extrarowheight}{-3pt}
\label{table-scales_result}
\begin{tabular*}{\tblwidth}{@{}LLLLLLLLLLLLLL@{}}
\toprule
Mode & Label & FoV & ViT & Dice ↑ & Jaccard ↑ & HD ↓ & Mode & Label & FoV & ViT & Dice ↑ & Jaccard ↑ & HD ↓ \\
\midrule
\multirow{42}{*}{Global} & \multirow{6}{*}{RV} & \multirow{3}{*}{3M} 
&  h & \underline{0.9157} & \underline{0.8452} & \underline{4.2075} 
& \multirow{24}{*}{Local} & \multirow{6}{*}{RV} & \multirow{3}{*}{3M} 
&  h & \underline{0.8701} & \underline{0.7825} & \underline{2.1536} \\
&&& l & 0.9126 & 0.8400 & 4.2208 &&&& l & 0.8613 & 0.7655 & 2.2648 \\
&&& b & 0.7614 & 0.6158 & 5.8264 &&&& b & 0.6643 & 0.5100 & 2.9949\\
\cmidrule(lr){3-7} \cmidrule(lr){10-14}
&& \multirow{3}{*}{6M} 
&  h & \underline{0.8898} & \underline{0.8015} & \underline{5.4137} 
&&& \multirow{3}{*}{6M} &  h & 0.7956 & 0.6760 & 3.1890 \\
&&& l & 0.8814 & 0.7889 & 5.5198 &&&& l & \underline{0.8020} & \underline{0.6818} & \underline{3.2766} \\
&&& b & 0.6717 & 0.5065 & 8.0147 &&&& b & 0.5198 & 0.3687 & 4.8443\\
\cmidrule(lr){2-7} \cmidrule(lr){9-14}
& \multirow{6}{*}{FAZ} & \multirow{3}{*}{3M} 
 &  h & \underline{0.9824} & \underline{0.9659} & \underline{2.5791} &
 & \multirow{6}{*}{Artery} & \multirow{3}{*}{3M} 
&  h & 0.8580 & 0.7603 & 2.4182 \\
&&& l & 0.9790 & 0.9596 & 2.7280 &&&& l & \underline{0.8623} & \underline{0.7646} & \underline{2.3465} \\
&&& b & 0.9572 & 0.9189 & 3.3391 &&&& b & 0.6415 & 0.4848 & 3.2004 \\
\cmidrule(lr){3-7} \cmidrule(lr){10-14}
&& \multirow{3}{*}{6M} 
 &  h & \underline{0.9126} & \underline{0.8547} & \underline{3.0198}
&&& \multirow{3}{*}{6M} &  h & \underline{0.7996} & \underline{0.6804} & \underline{3.3929} \\
&&& l & 0.8935 & 0.8451 & 3.1292 &&&& l & 0.7902 & 0.6656 & 3.5075 \\
&&& b & 0.8666 & 0.7764 & 3.5500 &&&& b & 0.5040 & 0.3542 & 5.0931 \\
\cmidrule(lr){2-7} \cmidrule(lr){9-14}
& \multirow{6}{*}{Capillary} & \multirow{3}{*}{3M} 
 &  h & \underline{0.8755} & \underline{0.7791} & \underline{7.6283} 
 && \multirow{6}{*}{Vein} & \multirow{3}{*}{3M} 
 & h & \underline{0.8636} & \underline{0.7696} & \underline{1.9788} \\
&&& l & 0.8567 & 0.7498 & 7.9025 &&&& l & 0.8595 & 0.7624 & 2.0351\\
&&& b & 0.6166 & 0.4464 & 12.212 &&&& b & 0.6477 & 0.4908 & 2.6357\\
\cmidrule(lr){3-7} \cmidrule(lr){10-14}
&& \multirow{3}{*}{6M} 
 &  h & \underline{0.8090} & \underline{0.6807} & \underline{9.5887} 
 &&& \multirow{3}{*}{6M} & h & \underline{0.8129} & \underline{0.6950} & \underline{2.7561} \\
&&& l & 0.8051 & 0.6750 & 9.6237 &&&& l & 0.8058 & 0.6842 & 2.8651\\
&&& b & 0.5786 & 0.4087 & 13.745 &&&& b & 0.5465 & 0.3854 & 3.9116\\
\cmidrule(lr){2-7} \cmidrule(lr){8-14}
& \multirow{6}{*}{Artery} & \multirow{3}{*}{3M} 
&  h & \underline{0.8895} & \underline{0.8020} & \underline{4.0860} \\
&&& l & 0.8871 & 0.7980 & 4.0116 \\
&&& b & 0.7101 & 0.5524 & 5.3278 \\
\cmidrule(lr){3-7}
&& \multirow{3}{*}{6M} 
 &  h & \underline{0.8580} & \underline{0.7530} & \underline{5.0640} \\
&&& l & 0.8520 & 0.7439 & 5.0795 \\
&&& b & 0.6149 & 0.4458 & 6.9462 \\
\cmidrule(lr){2-7}
& \multirow{6}{*}{Vein} & \multirow{3}{*}{3M} 
 &  h & \underline{0.8872} & \underline{0.7982} & \underline{3.5825} \\
&&& l & 0.8777 & 0.7832 & 3.6662 \\
&&& b & 0.6881 & 0.5255 & 4.8370 \\
\cmidrule(lr){3-7}
&& \multirow{3}{*}{6M} 
 &  h & \underline{0.8599} & \underline{0.7558} & \underline{4.9363} \\
&&& l & 0.8504 & 0.7416 & 5.1042 \\
&&& b & 0.6180 & 0.4491 & 6.6714 \\
\bottomrule
\end{tabular*}
\end{table*}

%% file: sections/5_discussion.tex
\section{Discussion}

\subsection{Effectiveness of Prompt Points}

Except for FAZ, prompt points offer little to no benefit for global segmentation results, even with adjustments to their generated quantity and positions. However, the segmentation effect is significantly improved with the increase in the number of prompt points in local mode. The speculated cause is that the segmentation of objectives in OCTA images does not necessitate supplementary information for guiding the segmentation process in global mode. The SAM, containing a pre-trained ViT as an image encoder, possesses a sufficiently strong capability for understanding image features. This suffices to handle various OCTA segmentation tasks, including the differentiation of vascular types in AV segmentation. For segmentation tasks in global mode, the optimal strategy is not to use any prompt points. This not only saves effort but also avoids a decrease in model performance due to the misuse of prompt points. 

Prompt points become effective in local segmentation reflected in the quantity, type, and position. Increasing the number of prompt points when they are few can significantly improve the segmentation effect. Moreover, the use of positive prompt points far outperforms negative prompt points. All types of special points can effectively extract the vessel stem and the intersection requires the least number of prompt points. Bifurcation points yield the best prompting effect, followed by endpoints, and intersections perform the worst. The reason is that the intersections may lead to a misjudgment of a vessel branch belonging to the artery or vein. The combination of bifurcations and endpoints further enhances the segmentation effectiveness of vessel branches. Other combinations of different types of prompt point shave minimal improvement on the segmentation effectiveness and require providing more prompt points. Therefore, in practical operations of local segmentation, the optimal prompting strategy is to prioritize giving bifurcations and endpoints and avoid giving prompt points at the intersections of arteries and veins. Then, supplementing more positive and negative prompt points based on the segmentation effect to get further improvement. The FAZ segmentation could be regarded as a kind of borderline case of local and global segmentation, and proper prompt points may optimize the boundary segmentation.


\subsection{Fine-tuning}

\begin{figure*}
    \centering
    \includegraphics[width=1\textwidth]{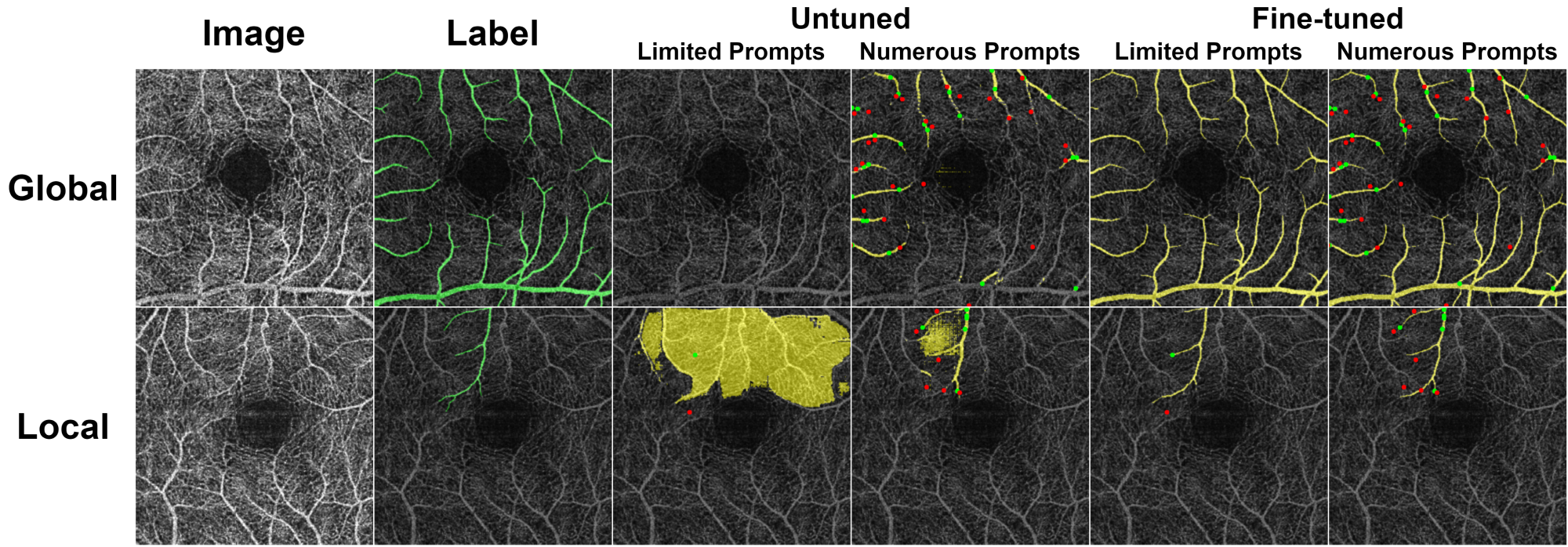}
    \caption{Comparison illustration of untuned and fine-tuned SAM-OCTA on RV segmentation.}
    \label{fig_result_fine-tuned}
\end{figure*}

Regardless of the ViT scale of the SAM model, the fine-tuning process is essential in both global and local modes. This not only clarifies the segmentation scope corresponding to the prompt points but also enhances the OCTA image feature comprehension of the image encoder. Figure \ref{fig_result_fine-tuned} displays the effect of fine-tuning on RV segmentation. In global mode, the SAM requires numerous prompt points for each vessel without fine-tuning. In this condition, segmented vessels often exhibit missed branches or fragmentation. Moreover, longer blood vessels are nearly impossible to segment effectively. In local segmentation,  the untuned SAM struggles to identify the target vessel to segment if there are only a few prompt points. With numerous prompt points, the model is able to focus on the target vessels effectively but still cannot segment accurately. For fine-tuned SAM-OCTA, the global segmentation does not require any prompt points. In local mode, only the least prompt points are needed to effectively segment the stem and most branches of the target vessels. Increasing the number of prompt points further enhances the segmentation performance of SAM-OCTA, enabling the segmentation of some minor branches.

The reason for the untuned model's performance is that the pre-trained SAM model is trained on a large dataset mainly consisting of internet images, which typically contain diverse object shapes with high contrast and may occupy a large region in the figure. For OCTA image segmentation, the model tends to perceive the segmentation target and its surrounding area as a whole. Fine-tuning aims to enable the image encoder to effectively extract shape features of vessels in OCTA images, including stem, branch morphology, and biomarkers such as diameter, length, curvature, and type.

\subsection{Limitation and Future Work}

Despite fully exploring the role of prompt points in SAM-OCTA through a series of experiments, this research still has two main limitations. Firstly, the role and mechanism of negative prompt points have not been clearly elucidated. In the experiments of Table \ref{table-nontarget_result}, we explored whether negative prompt points have an inhibitory effect on non-target vessels in segmentation, and the results indicate that negative prompt points do not have the expected effect. However, in some cases, we find that negative prompt points could suppress the segmentation of vessels surrounding them upon examining predicted samples. Currently, we have not found clear evidence of unique advantages in using negative prompt points, and further work is demanded to explore the advantages of negative prompt points in segmentation.

Furthermore, the prompt points generation strategy may establish closer connections with biomarkers in OCTA images. The types of known biomarkers in OCTA images are diverse. Prompt points generated based on biomarkers may further elucidate the model's interpretability when segmenting a specific sample. The above works could help SAM-OCTA achieve more accurate medical diagnoses and facilitate its practical use in real-world scenarios.

%% file: sections/6_conclusion.tex
\section{Conclusion}

This paper utilizes the LoRA technique and proposes a prompt points generation approach to fine-tune SAM for segmentation tasks based on OCTA datasets. Our method is named SAM-OCTA, which achieves or approaches the SoTA results in multiple OCTA image segmentation tasks, and a detailed analysis of the effect of prompt points in multiple tasks is conducted. This method performs well in conventional OCTA segmentation tasks (RV and FAZ) and yields favorable outcomes in segmenting capillary, artery, and vein targets.

As a fine-tuning strategy, SAM-OCTA freezes the original parameters of SAM, thus preserving its existing powerful image semantic comprehension capabilities. Fine-tuning is essential for further improvement in segmentation performance. Fine-tuning serves a dual purpose: firstly, it helps the image encoder adapt to the unique OCTA image type; secondly, it facilitates the integration of the image and the prompt point features.

A strategy based on target annotation and connected components generation is adopted for prompt points. Label information is incorporated because if users provide prompt points when using relevant segmentation tools in practice, they will implicitly incorporate their perceived ground-truth expectation. The effect of prompt points varies for segmentation modes. For global mode, the prompt points have minimal contributions, but SAM-OCTA is still able to segment well. The effect of prompt points becomes significant in the exploratory local OCTA segmentation. Generally, the positive points have the primary role, and the performance improves as the number of points increases. We also experimented on the special points related to the biomarker. We find that the combination of bifurcations and endpoints is beneficial for vessel branch segmentation, while the intersection can ruin it.

From a practical perspective, we also conducted comparisons and analyses of the performance of models of different scales and the functionality before and after model fine-tuning. Our work aims to assist users in segmenting OCTA vascular and biomarker targets of their specific interest. SAM-OCTA is expected to be an effective tool for diagnosing OCTA-related diseases, especially those involving artery-vein lesion structures.